\documentclass{article}

\usepackage[nonatbib,final]{neurips_2020}

\usepackage[utf8]{inputenc} % allow utf-8 input
\usepackage[T1]{fontenc}    % use 8-bit T1 fonts
\usepackage{url}            % simple URL typesetting
\usepackage{nicefrac}       % compact symbols for 1/2, etc.

\usepackage[usenames,dvipsnames,svgnames,table]{xcolor}
\usepackage{xspace}
\usepackage{graphicx}

\usepackage{subcaption} % comment subfigure package while using it
\usepackage{microtype}
\usepackage{cite}
\usepackage{amsmath,amssymb}%,amsthm}
\usepackage{amsfonts}

\usepackage{enumitem} % to reduce space between items in enumerate

\usepackage[pagebackref=true,breaklinks=true,colorlinks,bookmarks=false]{hyperref}

\usepackage{algorithm}
\usepackage{algpseudocode}
\algrenewcommand\algorithmicrequire{\textbf{Input:}}
\algrenewcommand\algorithmicensure{\textbf{Output:}}
\usepackage{tabularx}
\makeatletter
\newcommand{\multiline}[1]{%
  \begin{tabularx}{\dimexpr\linewidth-\ALG@thistlm}[t]{@{}X@{}}
    #1
  \end{tabularx}
}
\makeatother

\def\onedot{\ifx\@let@token.\else.\null\fi\xspace}

\def\ie{\emph{i.e}\onedot}

\makeatother

\def\Vec#1{{\boldsymbol{#1}}} % vectors
\def\Mat#1{{\boldsymbol{#1}}} % martices
\def\GRASS#1#2{\mathcal{G}({#2},{#1})} % Grassmannain manifold
\def\ORTHO#1{\mathcal{O}({#1})} % the orthogonal group

\DeclareMathOperator{\Tr}{Tr}

\title{Affinity guided Geometric \\Semi-Supervised Metric Learning}

\author{Ujjal Kr Dutta,\textsuperscript{$\ddag$,$\dag$,1} Mehrtash Harandi,\textsuperscript{$\S$,2} C Chandra Sekhar\textsuperscript{$\dag$,3} 
\\
\textsuperscript{$\ddag$} Myntra Designs, 
\textsuperscript{$\dag$}
Indian Institute of Technology Madras, India, 
\textsuperscript{$\S$}
Monash University, Australia\\
{\small \textsuperscript{1}ukd@cse.iitm.ac.in, \textsuperscript{2}mehrtash.harandi@monash.edu, \textsuperscript{3}chandra@cse.iitm.ac.in}
}

\begin{document}

\maketitle

\setlength{\abovedisplayskip}{3pt}
\setlength{\belowdisplayskip}{3pt}

\begin{abstract}
In this paper, we revamp the forgotten classical Semi-Supervised Distance Metric Learning (SSDML) problem from a Riemannian geometric lens, to leverage stochastic optimization within a end-to-end deep framework. The motivation comes from the fact that apart from a few classical SSDML approaches learning a linear Mahalanobis metric, deep SSDML has not been studied. We first extend existing SSDML methods to their deep counterparts and then propose a new method to overcome their limitations. Due to the nature of constraints on our metric parameters, we leverage Riemannian optimization. Our deep SSDML method with a novel affinity propagation based triplet mining strategy outperforms its competitors.
\end{abstract}

%%%%%%%%% BODY TEXT
\section{Introduction: Extending Classical SSDML to Deep SSDML}
% In applications where it is infeasible to obtain class labels for all the examples, Semi-Supervised Distance Metric Learning (SSDML) approaches learn a discriminative (similar examples closer, and dissimilar examples further) distance metric using only a few labeled examples, while leveraging additionally available unlabeled data.
Distance Metric Learning (DML) learns a discriminative embedding by bringing similar examples closer, and moving away dissimilar ones. Semi-Supervised DML (SSDML) is useful in applications where we have class labels for only a few examples, with additionally available unlabeled data. Surprisingly, apart from a few classical SSDML approaches \cite{S3ML,LRML-b,SERAPH-b,ISDML,APSSML_ICANN18,Distributed_SERAPH_Access16} that learn a linear Mahalanobis metric, deep SSDML has not been studied (although deep semi-supervised methods for classification tasks \cite{oliver2018realistic,SSL_05,SSL_09,GAM_SSL_NeurIPS19,VAT_PAMI18,iscen2019label}, and \textit{supervised} deep DML methods \cite{tuplet_margin_ICCV19,multi_sim_CVPR19,SNR_CVPR19,circle_CVPR20,fastAP_CVPR19,proxyNCA_ICCV17,arcface_CVPR19,soft_triple_ICCV19} exist). We begin our paper with a straightforward extension of classical SSDML to deep SSDML.

Let $\mathcal{X}=\mathcal{X}_L \cup \mathcal{X}_U$ be a given dataset, consisting of a set of labeled examples $\mathcal{X}_L=\{\Vec{z}_i\in \mathbb{R}^d \}_{i=1}^{N_L}$ with the associated label set $\{y_i\}_{i=1}^{N_L}, (y_i\in\{1,\cdots,C\}; C \textrm{ is the number of classes})$, and a set of unlabeled examples $\mathcal{X}_U=\{\Vec{z}_j \in \mathbb{R}^d\}_{j=N_L+1}^{N}$. Existing SSDML approaches learn the parametric matrix $\Mat{M}\in \mathbb{R}^{d \times d}, \Mat{M} \succeq 0$ of the squared Mahalanobis distance metric $\delta^2_{\Mat{M}}(\Vec{z}_i,\Vec{z}_j)=(\Vec{z}_i-\Vec{z}_j)^\top \Mat{M} (\Vec{z}_i-\Vec{z}_j)$, for a pair of descriptors $\Vec{z}_i,\Vec{z}_j \in \mathbb{R}^d$ (classically obtained using hand-crafted features). The SSDML approaches can mainly be categorized under two major paradigms: i) entropy minimization \cite{entropy_minimization_NIPS05}, and ii) graph-based. The SERAPH \cite{SERAPH-b,Distributed_SERAPH_Access16} approach, the only representative from the first category, tries to maximize the entropy of a conditional probability distribution over the pairwise constraints obtained from labeled data, while minimizing the entropy over the unlabeled data. The other category, i.e., graph-based, consisting of LRML \cite{LRML-b}, ISDML \cite{ISDML}, APLLR \cite{APSSML_ICANN18}, and APIT \cite{APSSML_ICANN18}, involves variation of a \textit{Laplacian regularizer} \cite{LPP} that preserves the pairwise distances among all the examples in $\mathcal{X}$ (hence, unlabeled data as well). In addition, each of them has a supervised term leveraging pairwise constraints obtained from the limited labeled data.% (details in Appendix).
\begin{figure}[!htb]
%\vspace{-0.5cm}
\centering
\begin{subfigure}{0.47\columnwidth}
    	\centering
		\includegraphics[width=\columnwidth]{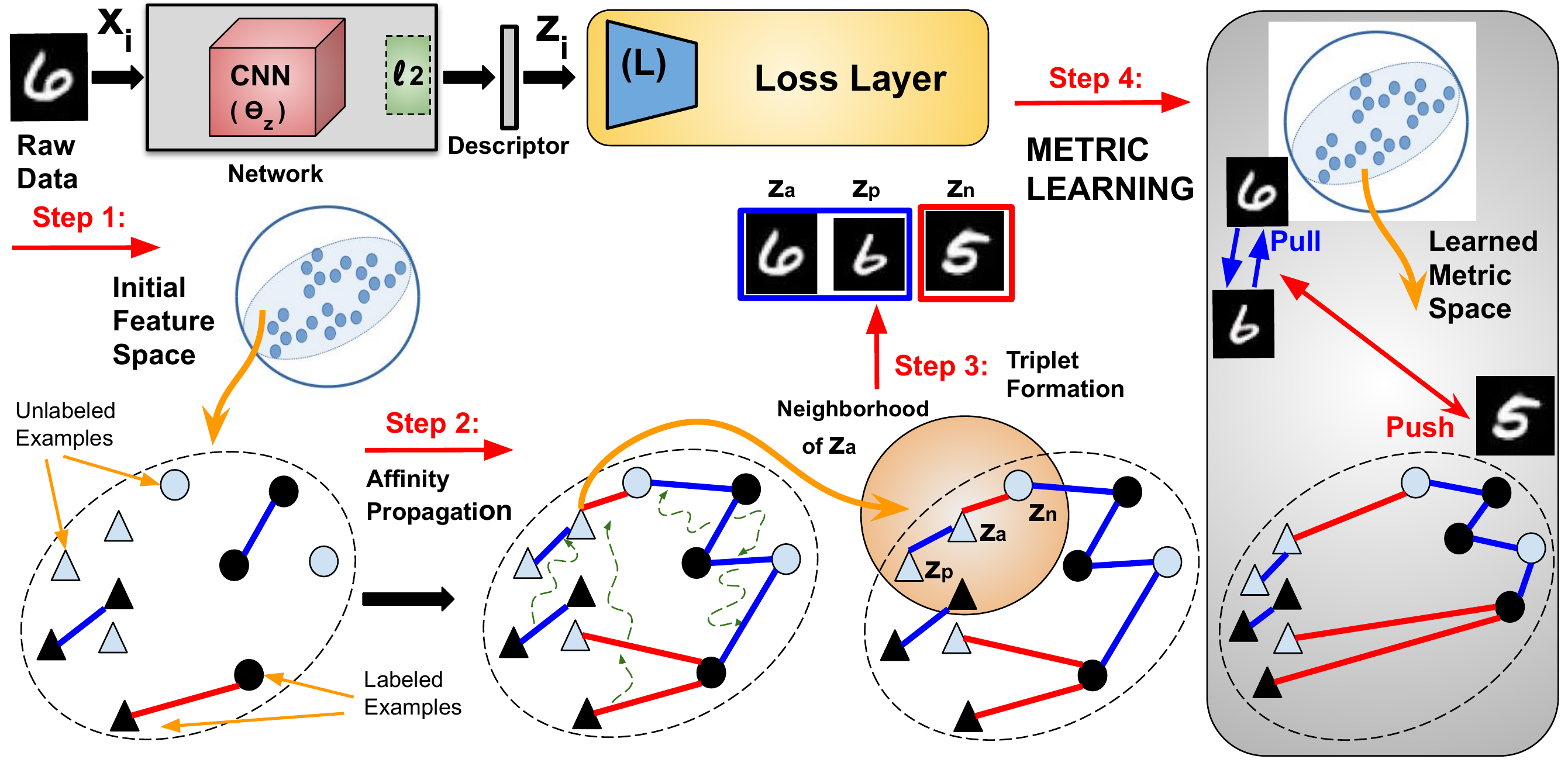}
		\caption{}
        \label{framework_agml}
\end{subfigure}
\hspace{0.2cm}
\begin{subfigure}{0.37\columnwidth}
    	\centering
		\includegraphics[width=\columnwidth]{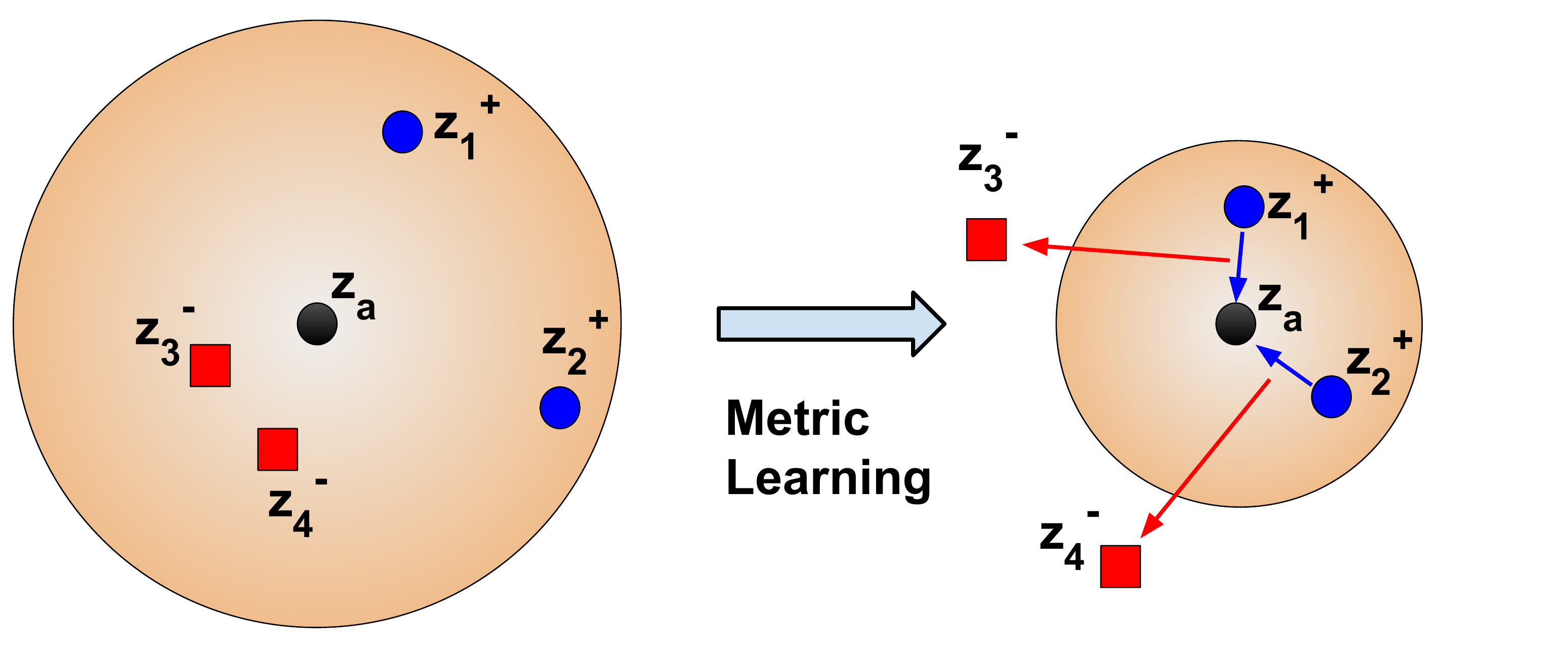}
		\caption{}
        \label{neigh_trip_mine}
\end{subfigure}
\caption{(a) A schematic illustration providing the intuition of our method (Best viewed in color). The raw image belongs to the MNIST dataset \cite{MNIST}, (b) Neighborhood triplet mining using propagated affinities (Best viewed in color). $\Vec{z}_a,\Vec{z}_1^+,\Vec{z}_2^+,\Vec{z}_3^-,\Vec{z}_4^-\in \mathcal{X}_L \cup \mathcal{X}_U^{(p)}$. An anchor $\Vec{z}_a$ (shown in black) is a point in the dataset that is currently in consideration for triplet mining within its $k$- neighborhood $\mathcal{N}_k(\Vec{z}_a)$, $k$= 4 (based on Euclidean distances in current embedding space). Points in blue ($\Vec{z}_1^+,\Vec{z}_2^+$) are more \textit{semantically} similar (by virtue of \textit{propagated affinities}) to the anchor, than the points in red ($\Vec{z}_3^-,\Vec{z}_4^-$). In the current embedding, the blue points are farther from the anchor $\Vec{z}_a$ than the red ones, despite having more semantic similarity. Hence, they should be pulled closer to the anchor in the learned space, compared to the red ones. }
\label{framework_mining}
\vspace{-0.6cm}
\end{figure}
\begin{figure}[!htb]
  \centering
\begin{minipage}[h]{0.3\textwidth}
\begin{algorithm}[H]
\begin{scriptsize}
\caption{Our stochastic approach}
\label{alg_AGML}
\begin{algorithmic}[1]
\State Initialize $\theta_z^0, \Mat{L}_0$.
\For {$t \gets 1$ to $T$}
\State \multiline{Fix $\theta_z^{t-1}$ and learn $\Mat{L}_t$. \label{update_L}}
\State \multiline{Fix $\Mat{L}_t$ and learn $\theta_z^t$. \label{update_th}}
\EndFor\label{for_count}%\textcolor{blue}{\Comment{End of count for loop.}}
\State \textbf{return} $\theta_z^T,\Mat{L}_T$
%\EndProcedure
\end{algorithmic}
\end{scriptsize}
\end{algorithm}
\end{minipage}
\hspace{0.2cm}
\begin{minipage}[h]{0.65\textwidth}
    \centering
    %\raisebox{-\height}{\includegraphics[width=1\textwidth]{xx.eps}}
    \includegraphics[width=0.5\linewidth]{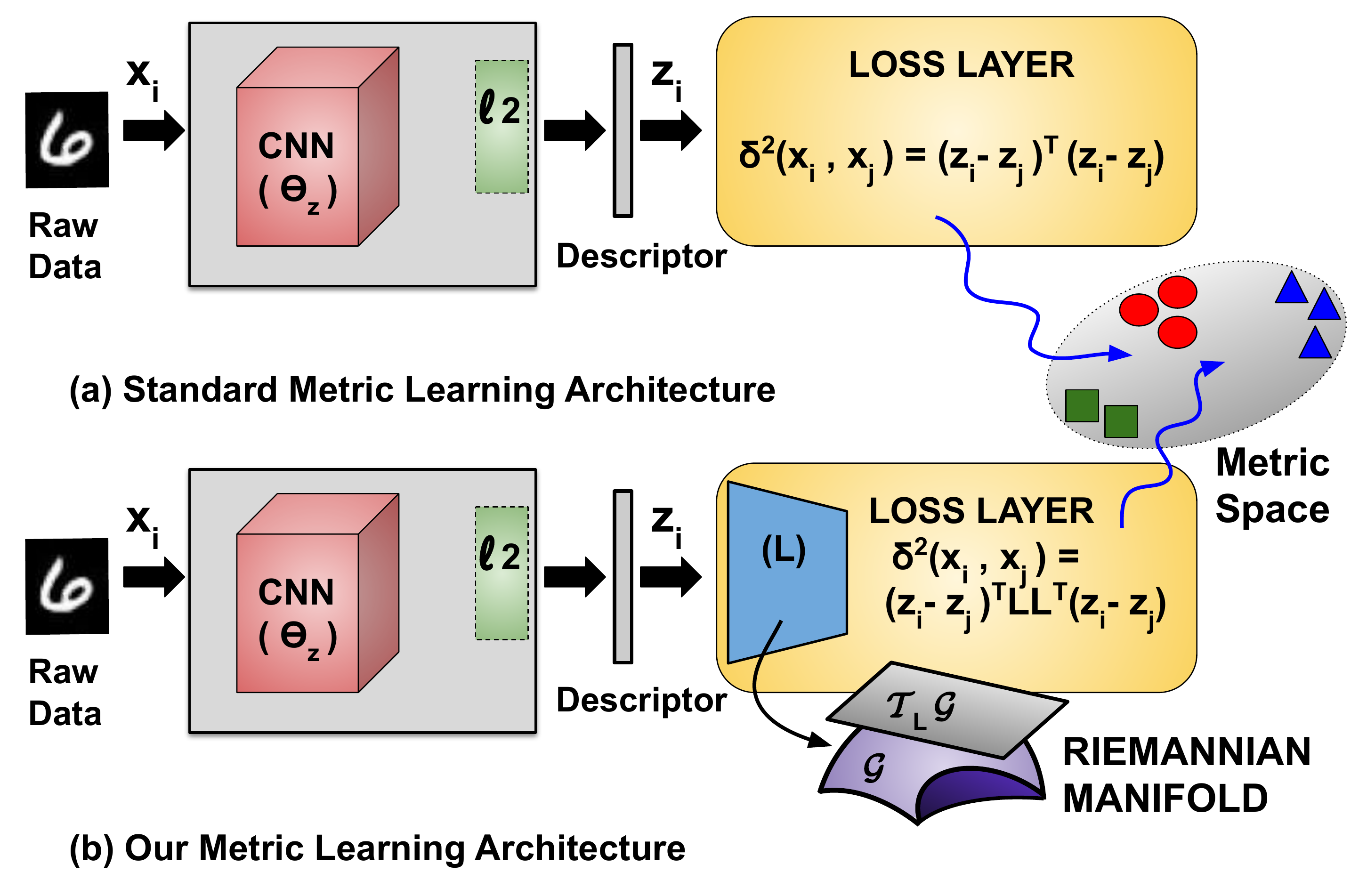}
    \caption{Our architecture to facilitate Riemannian optimization.}
    \label{agml_architecture}
    \vspace{-1cm}
\end{minipage}
\end{figure}

As the losses in these methods are differentiable, we can backpropagate their gradients while using mini-batch based stochastic optimization. For the graph-based methods, one could first sample a random partition of the unlabeled examples, and construct a sub-graph along with the limited number of available labeled examples. Then, the mini-batches can be drawn from this partition alone for several epochs (over this partition). This process could be iterated over several partitions over the entire dataset. The same partition based strategy could be used for SERAPH as well, with the only exception that it does not require graph construction. Formally, we can learn a deep embedding that induces a distance metric of the form $\delta^2_{\Mat{M}}(\Vec{z}_i,\Vec{z}_j)$, such that the descriptors $\Vec{z}_i,\Vec{z}_j \in \mathbb{R}^d$ are obtained using a deep neural network $z:\mathcal{X}\rightarrow \mathbb{R}^d$ with parameters $\theta_z$, while simultaneously learning $(\theta_z,\Mat{M})$ in an end-to-end manner. However, in many cases, the full-rank matrix $\Mat{M}$ of the distance metric learns redundancies in the data, which degrades the quality of the nonlinear embedding. Hence, we utilize the property that $\Mat{M} \succeq 0$, and factorize it as: $\Mat{M}=\Mat{L}\Mat{L}^\top$ s.t. $\Mat{L} \in \mathbb{R}^{d \times l}, l \leq d$, and rather propose to learn the metric wrt the matrix $\Mat{L}$ ($O(dl)$ parameters).% (Algorithm \ref{alg_AGML}).

\section{Proposed approach}
Because of a few limitations of the existing SSDML methods (discussed later in the Appendix), we now propose a method for SSDML, which is illustrated in Figure \ref{framework_agml}. 
% \begin{figure}[t]
% \centering
% 	\includegraphics[width=0.6\columnwidth]{AGML_framework.pdf}
%     \caption{A schematic illustration providing the intuition of our method (Best viewed in color). The raw image belongs to the MNIST dataset \cite{MNIST} }
%     \label{framework_agml}
% \vspace{-8mm}%Put here to reduce too much white space after your table
% \end{figure}
In the semi-supervised DML setting, the first task is to \textit{mine} informative constraints, and then use an appropriate loss function. As the first stage, we propose a novel method for mining constraints using a graph-based technique while leveraging affinity propagation \cite{S3ML}. Let $\mathcal{X}_U^{(p)}$ be a randomly selected partition of unlabeled data. We construct a kNN graph using $\mathcal{X}_L \cup \mathcal{X}_U^{(p)}$, s.t. the nodes represent the examples, and edge weights denote the \textit{affinities} (or similarities) among the examples. The initial affinity matrix $\Mat{W}^0\in\mathbb{R}^{(N_L+N_p)\times (N_L+N_p)}$ is defined as follows: i) $W_{ij}^0 = +1 \mbox{, if } i \neq j \textrm{ and } \exists y_i, y_j, \textrm{ s.t. } y_i = y_j$, ii) $W_{ij}^0 = -1 \mbox{, if } i \neq j \textrm{ and } \exists y_i, y_j, \textrm{ s.t. } y_i \neq y_j$, iii) $W_{ij}^0 = +1 \mbox{, if } i = j$, and iv) $W_{ij}^0 =0 \mbox{} \textrm{, otherwise}$. Here, $N_L$ and $N_p$ are respective cardinalities of $\mathcal{X}_L$ and $\mathcal{X}_U^{(p)}$.

The neighborhood structure of the kNN graph can be indicated using a matrix $\Mat{Q} \in\mathbb{R}^{(N_L+N_p)\times (N_L+N_p)}$ defined as: i) $Q_{ij} = 1/k \mbox{, if } \Vec{z}_j \in \mathcal{N}_k(\Vec{z}_i)$, and ii) $Q_{ij} = 0 \mbox{} \textrm{, otherwise}$. Here, $\mathcal{N}_k(\Vec{z}_i)$ is the set of $k$-nearest neighbor examples of $\Vec{z}_i$. The flow of information from the non-zero entries of $\Mat{W}^0$ (representing the labeled pairs) to the zero entries (representing the unlabeled pairs), can be performed by using a closed-form expression described by Liu \textit{et al.} \cite{S3ML}, as: $\Mat{W}^*=(1-\gamma)(\mathbf{I}_{N_L+N_p}-\gamma \Mat{Q})^{-1}\Mat{W}^0$. This step is called \textit{affinity propagation}. This step essentially performs a random-walk to propagate the affinities, where $0<\gamma<1$ is a weight hyper-parameter between the affinities obtained at the current step to that of the initial one. As $N_L+N_p\ll N$, we do not encounter any difficulty while scaling up to large datasets. To obtain a symmetric affinity matrix $\Mat{W}$, we perform a final symmetrization step as follows: $W_{ij}=(W^*_{ij}+W^*_{ji})/2$.

The finally obtained representation of the symmetric affinity matrix $\Mat{W}$ is used to mine triplets for metric learning. In doing so, we take into account the $k$-neighborhood $\mathcal{N}_k(\Vec{z}_a)$ of an example $\Vec{z}_a\in \mathcal{X}_L \cup \mathcal{X}_U^{(p)}$ that we consider as an anchor. Let $\mathcal{N}_W(\Vec{z}_a)=\{\Vec{z}^+_1,\Vec{z}^+_2,\cdots,\Vec{z}^+_{k/2},\Vec{z}^-_{k/2+1},\Vec{z}^-_{k/2+2},\cdots,\Vec{z}^-_{k}\}$ be the $k$-neighboring examples of $\Vec{z}_a$ sorted in descending order of their \textit{propagated affinities} w.r.t. $\Vec{z}_a$, i.e., $\Mat{W}(\Vec{z}_a,\Vec{z}^+_1)>\cdots>\Mat{W}(\Vec{z}_a,\Vec{z}^+_{k/2})>\Mat{W}(\Vec{z}_a,\Vec{z}^-_{k/2+1})>\cdots>\Mat{W}(\Vec{z}_a,\Vec{z}^-_k)$. Essentially, $\mathcal{N}_W(\Vec{z}_a)$ is simply a sorted version of $\mathcal{N}_k(\Vec{z}_a)$. As the obtained affinities are an indication of the semantic similarities among examples, we take it as a guidance to form triplets. Given an anchor $\Vec{z}_a$, intuitively we can consider an example $\Vec{z}^+_i$ with more affinity towards $\Vec{z}_a$ as a \textit{positive}, and another example $\Vec{z}^-_j$ with lesser affinity towards $\Vec{z}_a$ as a \textit{negative}, and form a triplet $(\Vec{z}_a,\Vec{z}^+_i,\Vec{z}^-_j)$. By considering first half of examples in the sorted neighborhood $\mathcal{N}_W(\Vec{z}_a)$ as \textit{positives}, and remaining half as \textit{negatives}, we can form the following triplets: $(\Vec{z}_a,\Vec{z}^+_1,\Vec{z}^-_{k/2+1}), (\Vec{z}_a,\Vec{z}^+_2,\Vec{z}^-_{k/2+2}),\cdots, (\Vec{z}_a,\Vec{z}^+_{k/2},\Vec{z}^-_{k})$. One may select the set of anchors from entire $\mathcal{X}_L \cup \mathcal{X}_U^{(p)}$, or by seeking the modes of the graph, without loss of generality. Figure \ref{neigh_trip_mine} illustrates our proposed novel idea of affinity-based triplet mining.
% \begin{figure}[t]
% \centering
% 	\includegraphics[width=0.7\linewidth]{neighborhood_triplet_mining.pdf}
%     \caption{Neighborhood triplet mining using propagated affinities (Best viewed in color). $\Vec{z}_a,\Vec{z}_1^+,\Vec{z}_2^+,\Vec{z}_3^-,\Vec{z}_4^-\in \mathcal{X}_L \cup \mathcal{X}_U^{(p)}$. An anchor $\Vec{z}_a$ (shown in black) is a point in the dataset that is currently in consideration for triplet mining within its $k$- neighborhood $\mathcal{N}_k(\Vec{z}_a)$, $k$= 4 (based on Euclidean distances in current embedding space). Points in blue ($\Vec{z}_1^+,\Vec{z}_2^+$) are more \textit{semantically} similar (by virtue of \textit{propagated affinities}) to the anchor, than the points in red ($\Vec{z}_3^-,\Vec{z}_4^-$). In the current embedding, the blue points are farther from the anchor $\Vec{z}_a$ than the red ones, despite having more semantic similarity. Hence, they should be pulled closer to the anchor in the learned space, compared to the red ones.}
%     \label{neigh_trip_mine}
% \vspace{-0.8cm}%Put here to reduce too much white space after your table
% \end{figure}
Given $\mathcal{X}_L \cup \mathcal{X}_U^{(p)}$ and the corresponding $\Mat{W}$, assume that we have obtained a triplet set $\mathcal{T}_p=\bigcup_b \mathcal{T}_p^{(b)}$. Here, $\mathcal{T}_p^{(b)}$ is a mini-batch of $T_b$ triplets, and $b \in [1, \cdots, \left \lfloor \frac{|\mathcal{T}_p|}{T_b}  \right \rfloor]$. Let, $\mathcal{T}_p^{(b)}=\{ (\Vec{z}_i,\Vec{z}_i^+,\Vec{z}_i^-) \}_{i=1}^{T_b}$. Then, we propose a smooth objective to learn the parameters ($\Mat{L},\theta_z$) of our deep metric as follows:
\begin{equation}
\label{opt_prob_AGML}
{\scriptsize
\begin{aligned}
\min_{\Mat{L},\theta_z} J_{metric}=\sum_{i=1}^{T_b}\textrm{log}(1+\textrm{exp}(m_{i})).
\end{aligned}
}
\end{equation}
Here, $m_{i}=\delta^2_{\Mat{L}}(\Vec{z}_i,\Vec{z}_i^+)-4\textrm{ tan}^2\alpha \textrm{ } \delta^2_{\Mat{L}}(\Vec{z}_i^-,\Vec{z}_{i-avg})$, s.t., $\Vec{z}_{i-avg}=(\Vec{z}_i+\Vec{z}_i^+)/2$, and $\delta^2_{\Mat{L}}(\Vec{z}_i,\Vec{z}_j)=(\Vec{z}_i-\Vec{z}_j)^\top\Mat{L}\Mat{L}^\top(\Vec{z}_i-\Vec{z}_j)$. $m_i$ tries to pull the anchor $\Vec{z}_i$ and the positive $\Vec{z}_i^+$ together, while moving away the negative $\Vec{z}_i^-$ from the mean $\Vec{z}_{i-avg}$, with respect to an angle $\alpha >0$ at the negative $\Vec{z}_i^-$ \cite{Angular_loss}. Given the fact that $J_{metric}$ is fully-differentiable, we can backpropagate the gradients to learn the parameters $\theta_z$ using SGD. This helps us in integrating our method within an end-to-end deep framework using Algorithm \ref{alg_AGML}, the theoretical analysis of which is present in the Appendix.

Additionally, we constrain $\Mat{L}\in \mathbb{R}^{d \times l}, l \leq d$ to be a orthogonal matrix, i.e., $\Mat{L}^\top\Mat{L}=\mathbf{I}_l$. This is because orthogonality acts as a regularizer to avoid degenerate embeddings due to a model collapse, as shown in our experiments. We conjecture that this is because \textit{orthogonality} omits redundant correlations among different dimensions, thereby omitting redundancy in the metric, which could otherwise hurt the generalization ability due to overfitting. This is in contrast to other regularizers that only constrain the values of the elements of a parametric matrix (like $l_1$ or $l_2$ regularizers). As $\Mat{L}$ with orthogonality naturally lies on a Stiefel manifold \cite{AMS09}, we make use of Riemannian geometry to impose the orthogonality constraint. It should be noted that for the orthogonal group $\ORTHO{l}=\{\Mat{B}\in \mathbb{R}^{l\times l}:\Mat{B}\Mat{B}^\top=\Mat{B}^\top\Mat{B}=\mathbf{I}_l\}$, and a matrix $\Mat{B}\in \ORTHO{l}$, replacing $\Mat{L}$ in $\delta^2_{\Mat{L}}(\Vec{z}_i,\Vec{z}_j)$ by $\Mat{L}\Mat{B}$ does not change the value of the objective, because $(\Mat{L}\Mat{B})(\Mat{L}\Mat{B})^\top=\Mat{L}\Mat{L}^\top$.

From a Riemannian geometric perspective \cite{AMS09}, we say that the objective is invariant to the right action of the group $\ORTHO{l}$. Hence, the correct geometry to consider for $\Mat{L}$ is that of the Grassmann manifold $\GRASS{d}{l}$ \cite{AMS09}. For each mini-batch of triplets $\mathcal{T}_p^{(b)}$, we perform Riemannian optimization to learn the parametric matrix $\Mat{L}$ while maintaining the constraint $\Mat{L} \in \GRASS{d}{l}$. After a few iterations of Riemannian optimization, we fix the value of the obtained $\Mat{L}$ and backpropagate the gradients for the rest of the network to learn $\theta_z$. We can repeat the process for several randomly sampled $\mathcal{X}_U^{(p)}$, until convergence. Note that we only have to perform Riemannian optimization in the loss layer. As such, we do not require any major changes to the standard deep metric learning architecture (as shown in Figure \ref{agml_architecture}).
% % \begin{figure}[t]
% % \centering
% % 	\includegraphics[width=\columnwidth]{figures/AGML_architecture.pdf}
% %     \caption{Comparison of our proposed architecture against standard metric learning architecture. The raw data image belongs to the MNIST dataset \cite{MNIST}}
% %     \label{agml_architecture}
% % %\vspace{-8mm}%Put here to reduce too much white space after your table
% % \end{figure}

\section{Empirical Observations}

\begin{table*}[!htb]
\vspace{-0.5cm}
\centering
\caption{Comparison against state-of-the-art SSDML approaches on MNIST, Fashion-MNIST and CIFAR-10.}
\label{ssdml_all_sota}
\resizebox{0.7\linewidth}{!}{%
\begin{tabular}{|c|ccccc|ccccc|ccccc|}
\hline
\textbf{Dataset} & \multicolumn{5}{c|}{\textbf{MNIST}}                                           & \multicolumn{5}{c|}{\textbf{Fashion-MNIST}}                                   & \multicolumn{5}{c|}{\textbf{CIFAR-10}}                                        \\ \hline
\textbf{Method}  & \textbf{NMI}  & \textbf{R@1}  & \textbf{R@2}  & \textbf{R@4}  & \textbf{R@8}  & \textbf{NMI}  & \textbf{R@1}  & \textbf{R@2}  & \textbf{R@4}  & \textbf{R@8}  & \textbf{NMI}  & \textbf{R@1}  & \textbf{R@2}  & \textbf{R@4}  & \textbf{R@8}  \\ \hline
Initial          & 17.4          & 86.5          & 92.4          & 95.9          & 97.6          & 32.6          & 71.9          & 82.0          & 89.5          & 94.8          & 20.4          & 38.2          & 54.9          & 71.0          & 83.6          \\ \hline \hline
Deep-LRML              & \textbf{48.3} & 88.7          & 93.2          & 95.9          & 97.6          & \textbf{53.2} & 75.3          & 83.9          & 90.2          & 95.0          & 16.0          & 40.2          & 56.3          & 71.5          & 83.7          \\
Deep-SERAPH            & 45.3          & 92.5          & 95.7          & 97.6          & 98.7          & 53.0          & 75.9          & 85.1          & 91.5          & 95.4          & 21.3          & 39.5          & 56.2          & 71.1          & 83.7          \\
Deep-ISDML             & 44.1          & 92.1          & 95.7          & 97.5          & 98.6          & 51.3          & 74.4          & 84.2          & 90.7          & 95.1          & 20.0          & 36.4          & 52.4          & 68.5          & 82.1          \\
Deep-APLLR             & 30.5          & 57.6          & 71.1          & 82.1          & 90.1          & 38.9          & 59.5          & 73.3          & 84.1          & 91.4          & 13.3          & 24.4          & 40.0          & 58.8          & 76.4          \\
Deep-APIT              & 31.7          & 87.6          & 92.9          & 96.0          & 97.9          & 37.3          & 69.4          & 80.1          & 88.5          & 94.1          & 11.4          & 26.0          & 41.4          & 59.6          & 76.4          \\ \hline
\textbf{Ours}    & 47.5          & \textbf{93.9} & \textbf{96.6} & \textbf{98.2} & \textbf{98.9} & 52.1          & \textbf{77.6} & \textbf{86.0} & \textbf{91.8} & \textbf{95.6} & \textbf{25.3} & \textbf{41.4} & \textbf{57.3} & \textbf{72.6} & \textbf{84.9} \\ \hline
\end{tabular}%
}
\vspace{-0.4cm}
\end{table*}

\begin{table}[!htb]
\vspace{-0.2cm}
\centering
\begin{minipage}[h]{0.45\linewidth}
% \usepackage{graphicx}
\centering
\caption{Comparison against state-of-the-art SSDML approaches on fine-grained datasets.}
\label{results_all_cub_cars}
\resizebox{\columnwidth}{!}{%
\begin{tabular}{|c|ccccc|ccccc|}
\hline
\textbf{Dataset}     & \multicolumn{5}{c|}{\textbf{CUB-200} }                                             & \multicolumn{5}{c|}{\textbf{Cars-196} }                                            \\ \hline \hline
\textbf{Method}      & \textbf{NMI}  & \textbf{R@1}  & \textbf{R@2}  & \textbf{R@4}  & \textbf{R@8}  & \textbf{NMI}  & \textbf{R@1}  & \textbf{R@2}  & \textbf{R@4}  & \textbf{R@8}  \\ \hline
Initial              & 34.3          & 31.7          & 42.2          & 55.4          & 67.9          & 23.7          & 24.2          & 32.8          & 43.4          & 54.9          \\ \hline \hline
Deep-LRML                  & 49.9          & 34.9          & 45.0          & 55.4          & 65.7          & 40.5          & 33.2          & 41.2          & 49.4          & 58.3          \\
Deep-SERAPH                & 50.5          & 39.7          & 52.2          & 64.4          & 76.5          & 37.2          & 34.8          & 46.7          & 59.4          & 71.7          \\
Deep-ISDML                 & 50.6          & 43.7          & 55.7          & 67.6          & 78.4          & 25.8          & 30.6          & 40.8          & 52.2          & 63.2          \\
Deep-APLLR                 & 38.9          & 25.7          & 36.6          & 48.7          & 61.9          & 33.2          & 33.0          & 43.6          & 55.6          & 67.9          \\
Deep-APIT                  & 52.1          & 42.2          & 54.5          & 66.7          & 78.2          & 37.3          & 38.5          & 50.7          & 62.7          & 74.9          \\ \hline \hline
\textbf{Ours} & \textbf{54.0} & \textbf{44.8} & \textbf{56.9} & \textbf{69.1} & \textbf{79.9} & \textbf{40.7} & \textbf{45.7} & \textbf{58.1} & \textbf{69.5} & \textbf{80.4} \\ \hline
\end{tabular}%
}
\vspace{-0.3cm}
\end{minipage}
\hspace{0.2cm}
\begin{minipage}[h]{0.45\linewidth}
% \usepackage{graphicx}
\centering
\caption{Comparison of our method against state-of-the-art deep unsupervised methods on fine-grained datasets.}
\label{cars_vs_unsup}
\resizebox{\columnwidth}{!}{%
\begin{tabular}{|c|ccccc|ccccc|}
\hline
\textbf{Dataset}     & \multicolumn{5}{c|}{\textbf{CUB-200} }                                             & \multicolumn{5}{c|}{\textbf{Cars-196} }                                            \\ \hline \hline
\textbf{Method}      & \textbf{NMI}  & \textbf{R@1}  & \textbf{R@2}  & \textbf{R@4}  & \textbf{R@8}  & \textbf{NMI}  & \textbf{R@1}  & \textbf{R@2}  & \textbf{R@4}  & \textbf{R@8}  \\ \hline
    Exemplar   & 45.0          & 38.2          & 50.3          & 62.8          & 75.0  & 35.4          & 36.5          & 48.1          & 59.2          & 71.0          \\
    Rotation   &49.1	&42.5	&55.8	&68.6	&79.4   &32.7	&33.3	&44.6	&56.4	&68.5          \\
    NCE       & 45.1          & 39.2          & 51.4          & 63.7          & 75.8            & 35.6          & 37.5          & 48.7          & 59.8          & 71.5          \\
    DeepCluster & 53.0          & 42.9          & 54.1          & 65.6          & 76.2& 38.5          & 32.6          & 43.8          & 57.0          & 69.5          \\ 
    Synthetic &53.4 &43.5 &56.2 &68.3 &79.1  &37.6 &42.0 &54.3 &66.0 &77.2          \\ \hline
    \textbf{Ours}        & \textbf{54.0} & \textbf{44.8} & \textbf{56.9} & \textbf{69.1} & \textbf{79.9}        & \textbf{40.7} & \textbf{45.7} & \textbf{58.1} & \textbf{69.5} & \textbf{80.4} \\ \hline
\end{tabular}%
}
\vspace{-0.3cm}

\end{minipage}
\end{table}

\begin{figure}[!htb]
\vspace{-0.1cm}
    \centering
    \includegraphics[trim={0cm 0cm 0cm 5.4cm},clip,width=0.4\columnwidth]{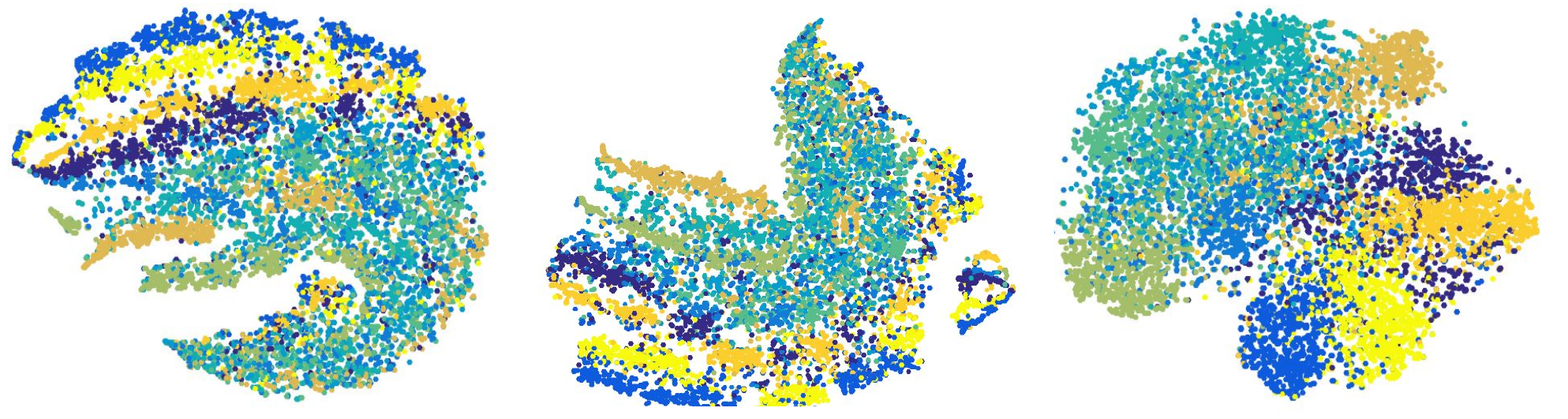}
    \caption{tSNE embeddings for the test examples of CIFAR-10. The left column represents the embeddings obtained right after random initialization. The embeddings obtained by our method: without orthogonality constraint on $\Mat{L}$ (middle column) and with orthogonality constraint on $\Mat{L}$ (rightmost column). Orthogonality leads to better, well-separated embeddings (see Table \ref{ablation_quant_orthog}). }
    \label{emb_AGML}
    \vspace{-0.1cm}
\end{figure}

\begin{table}[!htb]
\vspace{-0.5cm}
\centering
    %\caption{Global caption}
    \begin{minipage}{.45\linewidth}
      \caption{Quantitative comparison of the performance of our method, without (w/o orth) and with orthogonality (w/ orth) on the metric parameters.}
      \label{ablation_quant_orthog}
      \centering
        \resizebox{0.5\columnwidth}{!}{%
        \begin{tabular}{|c|ccccc|}
        \hline
        Method           & \textbf{NMI}  & \textbf{R@1}  & \textbf{R@2}  & \textbf{R@4}  & \textbf{R@8}  \\ \hline
        Dataset          & \multicolumn{5}{c|}{\textbf{MNIST}}                                           \\ \hline
        w/o orth         & 42.9          & 91.5          & 95.3          & 97.3          & 98.6          \\
        \textbf{w/ orth} & \textbf{47.5} & \textbf{93.9} & \textbf{96.6} & \textbf{98.2} & \textbf{98.9} \\ \hline
        Dataset          & \multicolumn{5}{c|}{\textbf{Fashion-MNIST}}                                   \\ \hline
        w/o orth         & 50.3          & 73.3          & 82.4          & 89.7          & 94.4          \\
        \textbf{w/ orth} & \textbf{52.1} & \textbf{77.6} & \textbf{86.0} & \textbf{91.8} & \textbf{95.6} \\ \hline
        Dataset          & \multicolumn{5}{c|}{\textbf{CIFAR-10}}                                        \\ \hline
        w/o orth         & 19.0          & 37.4          & 54.2          & 69.7          & 83.4          \\
        \textbf{w/ orth} & \textbf{25.3} & \textbf{41.4} & \textbf{57.3} & \textbf{72.6} & \textbf{84.9} \\ \hline
        \end{tabular}%
        }
    \end{minipage}%
    \hspace{0.2cm}
    \begin{minipage}{.45\linewidth}
      \centering
        \caption{Comparison of our method against supervised deep metric learning baselines on CUB.}
        \label{fgvc_vs_sup}
        \resizebox{0.5\columnwidth}{!}{%
        \begin{tabular}{|c|cccc|}
        \hline
        Method      & \textbf{R@1}  & \textbf{R@2}  & \textbf{R@4}  & \textbf{R@8}  \\ \hline
        Dataset     & \multicolumn{4}{c|}{\textbf{CUB-200}}                                         \\ \hline
        Triplet  &42.6 &55.2 &66.9 &77.6          \\ 
        Angular  &41.4	&54.5	&67.1	&78.4 \\ \hline
        \textbf{Ours}        & \textbf{44.8} & \textbf{56.9} & \textbf{69.1} & \textbf{79.9} \\ \hline
        
        % Dataset     & \multicolumn{5}{c|}{\textbf{Cars-196}}                                        \\ \hline
        
        % Exemplar \cite{ExemplarCNN_TPAMI16} (TPAMI-16)   & 35.4          & 36.5          & 48.1          & 59.2          & 71.0          \\
        % Rotation \cite{rotation_ICLR18} (ICLR-18)   &32.7	&33.3	&44.6	&56.4	&68.5          \\
        % NCE \cite{NCE_CVPR18} (CVPR-18)        & 35.6          & 37.5          & 48.7          & 59.8          & 71.5          \\
        % DeepCluster  \cite{DeepCluster_ECCV18} (ECCV-18) & 38.5          & 32.6          & 43.8          & 57.0          & 69.5          \\ 
        % Synthetic \cite{SUML_AAAI20} (AAAI-20) &37.6 &42.0 &54.3 &66.0 &77.2          \\ \hline
        % \textbf{Ours}        & \textbf{40.7} & \textbf{45.7} & \textbf{58.1} & \textbf{69.5} & \textbf{80.4} \\ \hline
        \end{tabular}%
        }
    \end{minipage} 
\vspace{-0.3cm}
\end{table}

In this section we evaluate our proposed method in terms of its effectiveness in clustering (wrt NMI) and retrieval (wrt R@K) tasks on a number of benchmark datasets (MNIST \cite{MNIST}, Fashion-MNIST \cite{Fashion-MNIST}, CIFAR-10\cite{CIFAR}, CUB-200 \cite{CUB} and Cars-196 \cite{Cars196}). We compare our method against 3 groups of approaches: i) Deep extensions of the discussed \textit{Semi-Supervised} DML methods (Deep-LRML \cite{LRML-b}, Deep-ISDML \cite{ISDML}, Deep-SERAPH \cite{SERAPH-b}, Deep-APLLR \cite{APSSML_ICANN18} and Deep-APIT \cite{APSSML_ICANN18}), ii) Recent state-of-the-art deep \textit{Unsupervised} representation learning methods (Exemplar \cite{ExemplarCNN_TPAMI16}, Rotation Net \cite{rotation_ICLR18}, NCE \cite{NCE_CVPR18}, DeepCluster \cite{DeepCluster_ECCV18} and Synthetic \cite{SUML_AAAI20}), and iii) \textit{Supervised} deep metric learning methods (Triplet \cite{FaceNet} and Angular \cite{Angular_loss}). We picked the Triplet \cite{FaceNet} and Angular \cite{Angular_loss} methods as they perform at par with more sophisticated state-of-the-art methods \cite{tuplet_margin_ICCV19,multi_sim_CVPR19,SNR_CVPR19,circle_CVPR20,fastAP_CVPR19,arcface_CVPR19,soft_triple_ICCV19}, when compared under a fair experimental protocol (as studied recently \cite{DML_reality_check_ECCV20,revisiting_DML_strategies_ICML20}). Details of experiments, performance metrics, datasets and baselines can be found in the Appendix.

Tables \ref{ssdml_all_sota}-\ref{results_all_cub_cars} show that we outperform the SSDML baselines (explanation of why our method outperforms these baselines is provided in the appendix). As shown in Table \ref{cars_vs_unsup}, better performance of our method in contrast to the unsupervised approaches demonstrates the benefit of additionally available labeled examples while learning a metric. Interestingly, many SSDML baselines perform poorer than unsupervised ones. This is because of the redundant information learned by their full-ranked matrices, thereby degrading the embedding information. However, this is not the case in our method. Lastly, as shown in Table \ref{fgvc_vs_sup}, better performance of our method in contrast to the fully-supervised baselines demonstrates the benefit of leveraging additional unlabeled examples via affinity propagation and our mining technique. On the other hand, the fully-supervised baselines overfit to the training data because of the availability of only a limited number of labeled examples. By using Table \ref{ablation_quant_orthog} and Figure \ref{emb_AGML}, we quantitatively and qualitatively demonstrate that orthogonality in the metric avoids degenerate embeddings. Orthogonality especially showed more benefit on the more complex CIFAR in comparison to the simpler MNIST dataset, particularly for the clustering task.

\section{Conclusions}
The main contributions of this paper are: i) A simple extension of classical linear approaches to address deep SSDML, which was not studied before. ii) To overcome limitations of existing SSDML methods, we propose a novel method that by virtue of its affinity propagation based novel triplet mining strategy outperforms its competitors. iii) In particular, we studied deep SSDML from a Riemannian geometric perspective, due to nature of our constraints. Without requiring any modifications in existing deep metric learning architecture, we facilitate Riemannian optimization within a deep stochastic framework.

% \section{Conclusions}
% In this paper we revisit and revamp the important problem of Semi-Supervised Distance Metric Learning (SSDML), in the end-to-end deep learning paradigm. We discuss a simple stochastic approach to extend classical SSDML methods to the deep SSDML setting. Owing to the theoretical limitations of the existing SSDML techniques, we propose a method to address these limitations. While the components in our method have been studied earlier, their composition has not been performed before. We show that by using this design composition, the overall approach could outperform existing approaches by the collective use of some of the best practices. In short, following are our major contributions: 1. Extension of classical SSDML approaches for end-to-end stochastic deep learning, with the proposal of a new method, 2. A novel triplet mining strategy leveraging graph-based affinity propagation, and 3. Adaptation of an angular variant of a metric learning loss with orthogonality imposed on the metric parameters to avoid a model collapse.

% \subsubsection*{Acknowledgements}
% We would like to thank the anonymous reviewers for their efforts at reviewing our work.

%\clearpage
{\small
\bibliographystyle{unsrt}
\bibliography{AGML_DiffGeo4DL_NeurIPS20_arXiv}
}

\newpage

\section*{Appendix: An efficient algorithm, computational complexity, and convergence analysis}
\textbf{An efficient matrix-based algorithm for our proposed approach:} We now provide an efficient matrix based implementation of the algorithm for our method. Given $\mathcal{T}_p^{(b)} =\{ (\Vec{z}_i,\Vec{z}_i^+,\Vec{z}_i^-) \}_{i=1}^{T_b}$, construct the following matrices: $\Mat{A}=[\Vec{a}_1 \cdots \Vec{a}_i \cdots \Vec{a}_{T_b}] \in \mathbb{R}^{d \times T_b}$, $\Mat{P}=[\Vec{p}_1 \cdots \Vec{p}_i \cdots \Vec{p}_{T_b}] \in \mathbb{R}^{d \times T_b}$ and $\Mat{Q}=[\Vec{q}_1 \cdots \Vec{q}_i \cdots \Vec{q}_{T_b}]$ $\in \mathbb{R}^{d \times T_b}$, where $\Vec{a}_i=(\Vec{z}_i+\Vec{z}_i^+)/2$, $\Vec{p}_i=\Vec{z}_i-\Vec{z}_i^+$ and $\Vec{q}_i= \Vec{z}_i^--\Vec{a}_i$. Let $\Vec{e}_l \in \mathbb{R}^l$ denote a vector containing 1 as all the components.
% Let $\boldsymbol \rho \in \mathbb{R}^{T_b}$ denote a vector, each component of which is computed as: $\rho_i=1+\exp(-\Vec{r}^T\Vec{c}_i)$. Then, we can compute the following vector: $\Vec{w}=\Vec{e}_{T_b} \oslash  \boldsymbol \rho$, where $\Vec{w} \in \mathbb{R}^{T_b}$ and $\oslash$ denotes the component-wise division operator.

We can compute a vector $\Vec{m}\in \mathbb{R}^{T_b}$ as: $\Vec{m}=(\Mat{L}^T\Mat{P} \odot \Mat{L}^T\Mat{P})^T\Vec{e}_l-4\textrm{ tan}^2\alpha(\Mat{L}^T\Mat{Q} \odot \Mat{L}^T\Mat{Q})^T\Vec{e}_l$, and $\Vec{f}\in \mathbb{R}^{T_b}$, such that each component is computed as: $f_i=\log(1+\exp(m_i))$, where $m_i$ denotes the $i$-th component of $\Vec{m}$. Then, the objective can be expressed as (assuming a fixed $\theta_z$): $J_{metric}=J(\Mat{L})=\Tr(\textrm{diag}(\Vec{f}))$. Here, $\Tr(.)$ is the trace operator and $\textrm{diag}(\Vec{f})$ is a diagonal matrix.

Now, let $\Vec{g}\in \mathbb{R}^{T_b}$ be a vector, components of which are computed as: $g_i=1/(1+\exp(-m_i))$. We can construct the following matrices: $\Mat{\tilde{P}}=[\Vec{\tilde{p}}_1 \cdots \Vec{\tilde{p}}_i \cdots \Vec{\tilde{p}}_{T_b}] \in \mathbb{R}^{d \times T_b}$ and $\Mat{\tilde{Q}}=[\Vec{\tilde{q}}_1 \cdots \Vec{\tilde{q}}_i \cdots \Vec{\tilde{q}}_{T_b}] \in \mathbb{R}^{d \times T_b}$ as follows: $\tilde{p}_{ik}=[2\Vec{g}^T]_ip_{ik}$, and $\tilde{q}_{ik}=[2\Vec{g}^T]_iq_{ik}$. Here, $p_{ik},\tilde{p}_{ik},q_{ik}$ and $\tilde{q}_{ik}$ are the $k$-th components of vectors $\Vec{p}_i,\Vec{\tilde{p}}_i,\Vec{q}_i$ and $\Vec{\tilde{q}}_i$ respectively. Also, $ [2\Vec{g}^T]_i$ denote the $i$-th component of the respective vector. Now, we can compute the Euclidean gradient as follows: $\nabla_\Mat{L}J(\Mat{L})=(\Mat{\tilde{P}}\Mat{P}^T-4\textrm{ tan}^2\alpha\Mat{\tilde{Q}}\Mat{Q}^T)\Mat{L}$.% Having the Euclidean gradient at our disposal, we can perform Riemannian Conjugate Gradient Descent (RCGD) to learn our parameters.

\textbf{Computational Complexity:} We now present the computational time complexity of the major steps involved: 
\textbf{i) Cost Function:} Computing the cost requires four matrix multiplications resulting in complexity of O($ldT_b$). Next, the transpose of the matrix products need to be computed, requiring O($lT_b$). Finally, the sum of losses across all triplets can be computed using a matrix trace operation requiring O($T_b$) complexity. \textbf{ii) Gradients:} The gradient with respect to $\Mat{L}$ requires the following computations: transposes requiring O($dT_b$), outer products requiring O($d^2T_b$). The subsequent products require O($d^2l$). Hence, the overall complexity is O($dT_b+d^2T_b+d^2l$).
% \textbf{iii) Parameter updates:} Major part of computation lies in the parameter update for $\Mat{L}$, which requires O($d^2l$). A final SVD requires O($d^2l+dl^2+l^3$). $l$ being lower, does not affect the complexity much. %Furthermore, the O($l^3$) complexity can be reduced to O($l^{2.376}$).
% \begin{itemize}[noitemsep,nolistsep]%topsep=0pt,itemsep=-1ex (OR) noitemsep,nolistsep
%     \item \textbf{Cost Function:} Computing the cost requires four matrix multiplications resulting in complexity of O($ldT_b$). Next, the transpose of the matrix products need to be computed, requiring O($lT_b$). Finally, the sum of losses across all triplets can be computed using a matrix trace operation requiring O($T_b$) complexity.
%     \item \textbf{Gradients:} The gradient with respect to $\Mat{L}$ requires the following computations: transposes requiring O($dT_b$), outer products requiring O($d^2T_b$). The subsequent products require O($d^2l$). Hence, the overall complexity is O($dT_b+d^2T_b+d^2l$).
%     \item \textbf{Parameter updates:} Major part of computation lies in the parameter update for $\Mat{L}$, which requires O($d^2l$). A final SVD requires O($d^2l+dl^2+l^3$). $l$ being lower, does not affect the complexity much.%Furthermore, the O($l^3$) complexity can be reduced to O($l^{2.376}$).
% \end{itemize}
Our proposed algorithm is linear in terms of the number of triplets in a mini-batch, i.e., $T_b$, which is usually low. The complexity of our algorithm is either linear or quadratic in terms of the original dimensionality $d$ (which in practice is easily controllable within a neural network).

It should be noted that in contrast to existing SSDML approaches, our method also enjoys a lower theoretical computational complexity. For example, LRML has a complexity of O($d^3$), while APLLR and APIT are of the order $O(d^3+N^3)$.

\paragraph{Convergence analysis}
Algorithm \ref{alg_AGML} provides a pseudo-code of our method to jointly learn $\theta_z$ and $\Mat{L}$ in an end-to-end manner. Here, $\theta_z^t$ and $\Mat{L}_t$ denote the values of the parameters at the $t^{th}$ iteration.
In line \ref{update_L} of Algorithm \ref{alg_AGML}, for a fixed $\theta_z^{t-1}$, we employ Riemannian optimization to learn $\Mat{L}_t$. It is done by computing the following Riemannian gradient: $\textrm{grad}_{\Mat{L}}= \nabla_{\Mat{L}}\mathcal{L}-\Mat{L}_{t-1}\Mat{L}_{t-1}^\top\nabla_{\Mat{L}}\mathcal{L}$. Here, let $\mathcal{L}(.,.)$ denotes the objective $J_{metric}$ and $\nabla_{\Mat{L}}\mathcal{L}$ denotes the Euclidean gradient computed as discussed earlier. The next iterate is obtained as: $\Mat{L}_t=\mathcal{R}_{\Mat{L}_{t-1}}(\xi_{\Mat{L}_{t-1}})$, where $\xi_{\Mat{L}_{t-1}}=-\eta \textrm{ grad}_{\Mat{L}}$ is a tangent vector pointing in the negative direction of the Riemannian gradient $\textrm{grad}_{\Mat{L}}$, and $\mathcal{R}(.)$ denotes the retraction. It is well known that moving along the negative direction of the Riemannian gradient minimizes the objective. Hence, we have $\boxed{ \mathcal{L}(\theta_z^{t-1},\Mat{L}_t) \leq \mathcal{L}(\theta_z^{t-1},\Mat{L}_{t-1}) }$. In line \ref{update_th} of Algorithm \ref{alg_AGML}, for a fixed $\Mat{L}_t$, we employ SGD to learn $\theta_z^t$. Therefore, theoretical guarantees of SGD can be directly applied, \ie, the last iterate can converge at the rate of $O(1/\sqrt{t})$ for updates of the form $\theta_z^t \gets \theta_z^{t-1} + \frac{1}{\sqrt{t}} \nabla_{\theta} \mathcal{L} (\theta_z^{t-1}, \Mat{L}_t)$. Thus we have $\boxed{ \mathcal{L}(\theta_z^t,\Mat{L}_t) \leq \mathcal{L}(\theta_z^{t-1},\Mat{L}_t) }$. Furthermore, due to the log-sum-exponential formulation, the objective is bounded, \ie, $\mathcal{L}(.,.) \geq \epsilon$, where $\epsilon \rightarrow 0, \epsilon>0$. Since $\mathcal{L}(\theta_z^{t-1},\Mat{L}_t) \leq \mathcal{L}(\theta_z^{t-1},\Mat{L}_{t-1})$ and $\mathcal{L}(\theta_z^t,\Mat{L}_t) \leq \mathcal{L}(\theta_z^{t-1},\Mat{L}_t)$ holds true for each iteration, and the objective is bounded, we have $\boxed{ \mathcal{L}(\theta_z^t,\Mat{L}_t) \leq \mathcal{L}(\theta_z^{t-1},\Mat{L}_{t-1}) }$, indicating that our method converges.% The empirical convergence of our method is illustrated in Figure \ref{ablation_alpha}.

\newpage
\section*{Appendix: Details of Datasets, Implementation, and Baseline Approaches}
\textbf{Datasets:} Following recent literature, the benchmark datasets that have been used are as follows:
%\begin{enumerate}[topsep=-1ex,itemsep=-1ex]%topsep=0pt,itemsep=-1ex (OR) noitemsep,nolistsep
\begin{itemize}[topsep=0pt,itemsep=0pt]%topsep=0pt,itemsep=-1ex (OR) noitemsep,nolistsep
    \item \textbf{MNIST} \cite{MNIST}: It is a benchmark dataset that consists of 70000 gray-scale images of handwritten digits. Each image is of $28\times 28$ pixels. There are 60000 training images and 10000 test images in the standard split.
    \item \textbf{Fashion-MNIST} \cite{Fashion-MNIST}: It is a similar dataset as the MNIST, but consists of images from 10 categories of fashion products. There are 60000 training images and 10000 test images in the standard split.
    \item \textbf{CIFAR-10} \cite{CIFAR}: This dataset consists of colour images of $32 \times 32$ pixels, containing animal or vehicle objects from 10 different categories. There are 50000 training images and 10000 test images.
    \item \textbf{CUB-200} \cite{CUB}: This dataset consists of images of 200 species of birds with first 100 species for training (5864 examples) and remaining for testing (5924 examples).
    \item \textbf{Cars-196} \cite{Cars196}: It consists of images of cars belonging to 196 models. The first 98 models containing 8054 images are used for training. The remaining 98 models containing 8131 images are used for testing.
\end{itemize}
The MNIST, Fashion-MNIST and CIFAR-10 datasets are widely used benchmarks with sufficiently large number of images for a comparative evaluation of different approaches. The CUB-200 and Cars-196 datasets are well-known for their use in \textit{Fine-Grained Visual Categorization} (FGVC), and have huge intra-class variances and inter-class similarities.

\textbf{Implementation details:} We adapted the network architectures in the MatConvNet tool \cite{MatConvNet}. For MNIST and Fashion datasets, the network for MNIST has been adapted as: Conv1($5\times5,20$) $\to$ max-pool $\to$ Conv2($5\times5,50$) $\to$ max-pool $\to$ Conv3($4\times4,500$) $\to$ ReLU $\to$ FC($500\times128$) $\to$ $l2$ $\to$ $\Mat{L}$($128\times64$). For CIFAR-10, we used the following adapted network: Conv1($5\times5,32$) $\to$ max-pool $\to$ ReLU $\to$ Conv2($5\times5,32$) $\to$ ReLU $\to$ avg-pool $\to$ Conv3($5\times5,64$) $\to$ ReLU $\to$ avg-pool $\to$ Conv4($4\times4,64$) $\to$ ReLU $\to$ $l2$ $\to$ $\Mat{L}$($64\times32$). For our method, we set $\gamma=0.99$ in the affinity propagation step, $k=10$ in the kNN graph, $\alpha=40^\circ$ in (\ref{opt_prob_AGML}), and initial learning rate $10^{-4}$. For MNIST and Fashion datasets, we choose 100 labeled examples (10 per class), while for CIFAR-10, we choose 1000 labeled examples (100 per class). We sample a random subset of 9k unlabeled examples and use it along with the labeled data to mine triplets. For each random subset, we run our method for 10 epochs (with mini-batch size of 100 triplets). In total, we run for a maximum of 50 epochs and choose the best model from across all runs. For MNIST and Fashion, we train upon randomly initialized networks. For CIFAR-10, we observed a better performance by pretraining with labeled examples (by replacing the $\Mat{L}$ \textit{layer} with softmax) for 30 epochs, and then fine-tune using our loss for 50 epochs. For all datasets, the graph has been constructed using the $l2$-normalized representations obtained just before $\Mat{L}$.

For the FGVC task, the GoogLeNet \cite{GoogLeNet} architecture pretrained on ImageNet \cite{ImageNet2015}, has been used as the backbone CNN, using MatConvNet \cite{MatConvNet}. We used the Regional Maximum Activation of Convolutions (R-MAC) \cite{RMAC_ICLR16} right before the average pool layer, aggregated over three input scales ($512$, $512/\sqrt{2}$, $256$). We choose five labeled examples per class. For our method, we set $\gamma=0.99$, $k=50$, $\alpha=45^\circ$ in (\ref{opt_prob_AGML}), and embedding size of 128. We take the entire dataset without partition based sampling and run for a maximum of 200 epochs (mini-batch size of 100 triplets), and choose the best model. Specifically, using our triplet mining strategy, we mined $146600$ triplets for the CUB dataset, and $201350$ triplets for the Cars dataset. In all experiments, we fix a validation dataset by sampling $15\%$ examples from each class of the training data. This validation dataset is used to tune the hyperparameters without taking any feedback from the test data. Note that to learn $\Mat{L}$ with orthogonal constraints we made use of the Manopt \cite{manopt} tool with CGD ($max\_iter=10$, with all other default settings).

\textbf{Compared state-of-the-art baseline approaches:}
We compare our proposed SSDML method against the following baseline techniques:
\begin{itemize}[topsep=0pt,itemsep=0pt]%topsep=0pt,itemsep=-1ex (OR) noitemsep,nolistsep
    \item \textbf{Deep-LRML}: This is the stochastic extension of the classical LRML method \cite{LRML-b} discussed earlier, by making use of our proposed stochastic approach. It follows the \textit{min-max principle} for the labeled data (minimizing distances between similar pairs, while maximizing distances between dissimilar pairs). A Laplacian regularizer is used to capture information from the unlabeled data.
    \item \textbf{Deep-ISDML}: Stochastic extension of the ISDML \cite{ISDML} method. It is similar to LRML, but makes use of densities around an example to adapt the Laplacian regularizer.
    \item \textbf{Deep-SERAPH}: Stochastic extension of the SERAPH \cite{SERAPH-b} method that makes use of the entropy minimization principle.
    \item \textbf{Deep-APLLR}: Stochastic extension of the APLLR \cite{APSSML_ICANN18} method. Makes use of a Log-Likelihood Ratio (LLR) based prior metric. The affinity propagation principle is used to propagate information from the labeled pairs to the unlabeled ones, and adapt the Laplacian (but no triplet mining like ours).
    \item \textbf{Deep-APIT}: Stochastic extension of the APIT \cite{APSSML_ICANN18} method. Makes use of an information-theoretic prior metric. The affinity propagation principle is used to adapt the Laplacian as in APLLR.
    \item \textbf{Exemplar} (TPAMI'16): This method \cite{ExemplarCNN_TPAMI16} attempts to learn the parameters associated with certain elementary transformation operations like translation, scaling, rotation, contrast, and colorization applied on random image patches.
    \item \textbf{Rotation Net} (ICLR'18): This method \cite{rotation_ICLR18} aims to learn representations of data that can accurately capture the information present in an image despite any rotation of the subject.
    \item \textbf{NCE} (CVPR'18): This method \cite{NCE_CVPR18} aims at bringing augmentations of an example together while moving away augmentations from different examples.
    \item \textbf{DeepCluster} (ECCV'18): This method \cite{DeepCluster_ECCV18} aims to jointly cluster metric representations while learning pseudo-labels for data in an end-to-end manner.
    \item \textbf{Synthetic} (AAAI'20): This method \cite{SUML_AAAI20} learns a metric using synthetic constraints generated in an adversarial manner.
    \item \textbf{Triplet} (CVPR'15): This method \cite{FaceNet} learns a metric using a standard triplet loss.
    \item \textbf{Angular} (ICCV'17): This method \cite{Angular_loss} learns a metric using an angular loss on a triplet of examples.
\end{itemize}

\textbf{Evaluation metrics:}
For comparing the approaches, we first learn a metric with the training data, and using it we obtain the test embeddings. The methods are compared based on their clustering (wrt NMI) and retrieval (wrt Recall@K, K=1,2,4,8) performances on the test embeddings. NMI is defined as the ratio of mutual information and the average entropy of clusters and entropy of actual ground truth class labels. The Recall@K metric gives us the percentage of test examples that have at least one K nearest neighbor from the same class. A higher value of all these metrics indicates a better performance for an approach.

\section*{Appendix: Why our method addresses the limitations of the existing SSDML methods, and performs better ?}
1. \textbf{LRML / ISDML vs Ours:} Both the LRML and ISDML methods define the affinity matrices directly using the distances among the initial representations. If the initial affinities are poor, the learned metric would be poor as well. On the other hand, our affinity matrix is adapted by the affinity propagation principle, while leveraging labeled data information as well. 2. \textbf{APLLR / APIT vs Ours:} Although APLLR and APIT make use of affinity propagation, they do not mine constraints using the enriched affinity matrix information, whereas we do. While their prior metric may be singular and hence poor, our method has no such dependency on a pre-computed metric. 3. \textbf{SERAPH vs Ours:} In contrast to SERAPH, our method has a stronger regularizer by virtue of orthogonality.

\end{document}